# Vedavani: A Benchmark Corpus for ASR on Vedic Sanskrit Poetry

**Sujeet Kumar[1,2], Pretam Ray[1], Abhinay Beerukuri[1], Shrey Kamoji[1],**

**Manoj Balaji Jagadeeshan[1], Pawan Goyal[1]**

[1]Indian Institute of Technology, Kharagpur
[2]B P Mandal College of Engineering, Madhepura
{ksujeet.cs, pretamroy309, beerukuriabhinay, shreykamoji, manojbalaji1}@gmail.com
pawang@cse.iitkgp.ac.in

## Abstract

Sanskrit, an ancient language with a rich linguistic heritage, presents unique challenges for automatic speech recognition (ASR) due to its phonemic complexity and the phonetic transformations that occur at word junctures, similar to the connected speech found in natural conversations. Due to these complexities, there has been limited exploration of ASR in Sanskrit, particularly in the context of its poetic verses, which are characterized by intricate prosodic and rhythmic patterns. This gap in research raises the question: *How can we develop an effective ASR system for Sanskrit, particularly one that captures the nuanced features of its poetic form?* In this study, we introduce Vedavani, the first comprehensive ASR study focused on Sanskrit Vedic poetry. We present a 54-hour Sanskrit ASR dataset, consisting of 30,779 labelled audio samples from the Rig Veda and Atharva Veda. This dataset captures the precise prosodic and rhythmic features that define the language. We also benchmark the dataset on various state-of-the-art multilingual speech models.[1] Experimentation revealed that IndicWhisper performed the best among the SOTA models.

## 1 Introduction

Sanskrit, an ancient and highly inflected language, holds great importance in preserving the knowledge of archaic India. Sanskrit is a language with fairly advanced disciplines of phonetics (Śikṣā), prosody (Chandas), and grammar (Vyākarana). The language has a rich oral tradition, and it tends to follow a phonemic orthography, resulting in systematic grapheme-phoneme correspondences. Connected speech leads to phonemic transformations in utterances, and in Sanskrit, this is faithfully preserved in writing as well (Krishna et al., 2018).

Recent advancements in Sanskrit automatic speech recognition (ASR) include the introduction of a corpus comprising 1,360 sentences (Anoop and Ramakrishnan, 2019). This was followed by the development of an end-to-end ASR system utilizing connectionist temporal classification (CTC), which demonstrated promising results with 5.5 hours of speech data (Suhani et al., 2023). Subsequently, the 78-hour Vakysancayah dataset was used to investigate various methods in acoustic modeling and language processing (Adiga et al., 2021). More recent contributions to the field include the Shrutilipi corpus with 27 hours of data (Javed et al., 2023) and the Kathbath dataset from the IndicSUPERB benchmark, which includes 102 hours of audio (Bhogale et al., 2023a).

However, despite these advances, there is still a significant gap. Existing datasets predominantly focus on prose, limiting their ability to capture the full linguistic diversity of Sanskrit. Much of the pre-classical and classical literature of the language is composed in verse, where the ordering of words is often dictated by metrical constraints rather than syntactic rules (Krishna et al., 2021) (Wright, 1968). This arbitrarily ordered word poses a unique challenge for ASR systems. To investigate whether current Sanskrit ASR models are capable of capturing the nuances of poetry, we evaluated the performance of four Sanskrit ASR models: SPRING-INX-MMS,[2] SPRING-INX-wav2vec2,[2] and SPRING-INX-ccc-wav2vec2,[2] along with IndicWhisper(Radford et al., 2023). The models depicted in Table 1 exhibit a poor performance, with Word Error Rates (WER) ranging from 99 to 110 and Character Error Rates (CER) from 39 to 46, underscoring the need for specialized datasets.

---

[1]Our dataset and code are publicly available at https://github.com/SujeetNlp/Vedavani

Table 1: Zero-shot inference results of existing Sanskrit ASR models on the poetry test dataset

| Models | WER | CER |
|---|---|---|
| SPG-INXS-MMS | 99.84 | 39.38 |
| SPG-INXS-CCC-W2V | 99.84 | 39.39 |
| SPG-INXS-W2V | 103.20 | 40.05 |
| IndicWhisper | 109.80 | 46.05 |

We, thus, introduce a new dataset, *Vedavani*, devoted to Vedic Sanskrit poetry. This collection consists of more than 54 hours of audio recordings totaling 30,779 sentences from the Atharva Veda (2000-1500 BCE) and Rig Veda (pre-2000 BCE) (Lal, 2015). *Vedavani* presents a special chance to learn the grammatical and stylistic subtleties of classical Sanskrit poetry since it catches the melodic rhythms, complex meters, and lyrical beauty of Vedic hymns and poems.

## 2 Data Collection

We curated textual resources for the Atharva Veda and Rig Veda from Wikipedia. Additionally, we sourced audio transcriptions of Vedic recitations from the Internet Archive, selecting recordings that accurately represent the traditional chanting style. We manually aligned the text and audio modalities with meticulous attention to create a well-structured dataset. This process involved segmenting the audio recordings into meaningful units corresponding to verses or phrases in the textual transcripts. Each audio segment was carefully mapped to its respective textual counterpart, ensuring synchronization between spoken and written recitations. The following sections provide a detailed explanation of our data alignment methodology, the challenges encountered, and the approaches used to ensure high-quality synchronization of textual and audio data.

### 2.1 Textual Data Extraction

We obtained textual content for the 20 kandas[3] of the Atharva Veda[4] and 10 mandalas[3] of the Rig Veda[5] from Wikipedia. The extraction process involved navigating the HTML structure to isolate and capture relevant text while excluding non-essential elements such as numbers, captions under images, etc. We ensured the text was free of special characters and standardized formatting for consistency, preserving spaces between lines and punctuation marks to maintain readability. This categorization facilitated easy retrieval of specific sections and optimized the dataset for further analysis and study.

---

[2] https://asr.iitm.ac.in/models

[3] Kandas and Mandalas refer to chapters in Atharva Veda and Rig Veda respectively

[4] https://sa.wikisource.org/wiki/अथर्ववेदः

[5] https://sa.wikisource.org/wiki/ऋग्वेदः

## 2.2 Audio Data Extraction

The audio recordings of the Atharva Veda[6] and Rig Veda[7] were obtained from the Internet Archive. Each recording is about 40–45 minutes long, totaling 27 files for the Atharva Veda and 54 files for the Rig Veda, with a combined size of 5.4GB. We converted these files to *wav* format and segmented them using pydub[8] library.

## 2.3 Textual-Audio Data Alignment

Upon segmenting the audio files into smaller chunks and obtaining corresponding text files, our expectation was for the audio files to align line by line with the Mandalas and Kandas of the Vedas, but we observed inconsistencies in the segmentation process, with audio chunks sometimes spanning multiple lines or only capturing part of a line from the Vedic texts. To resolve these issues, we adjusted the segmentation parameters, such as the minimum length of silence and the silence threshold. Specifically, the minimum length of silence determines the duration of silence (in milliseconds) required to trigger a split in the audio file, while the silence threshold specifies the decibel level below which audio is considered silent. We set the minimum length of silence in the range of 3-7 milliseconds and the silence threshold parameter to be in the range of -30 to -70 dB. Even after trying, all these inconsistencies still persisted. Hence, the decision was taken to manually align each of the chunks with its corresponding transcript.

During the manual alignment process, we added 500 ms of silence at the beginning and end of each audio file for aesthetic purposes, as it was difficult to comprehend the abrupt beginning and end of the audio against its respective transcript. For each of the audio files, we have manually listened to the audio and aligned its transcript from the text corpus. We removed segments shorter than 0.25 seconds of audio (1.25 seconds including silence). This process resulted in a set of organized audio files along with their transcripts suitable for analysis. With the textual and audio data aligned, we assembled what we now refer to as the *Vedavani Corpus*—a comprehensive dataset designed to facilitate model training and evaluation.

Table 2: Overview of the Sanskrit Vedic corpus, including data statistics and the train-validation-test split.

| Corpus | Number of Verses |
|---|---|
| **Total** | **30,779** |
| Rig Veda | 20,782 |
| Atharva Veda | 9,997 |
| **Train/Val/Test Split** | **Percentage** |
| Train | 24,623 (80%) |
| Validation | 3,078 (10%) |
| Test | 3,078 (10%) |

The dataset comprises a total of 30,779 verses, including 20,782 verses from the Rig Veda and 9,997 verses from the Atharva Veda. To facilitate model training and evaluation, the data is split into training, validation, and test sets, with proportions of 80%, 10%, and 10%, respectively. The dataset statistics are depicted in Table 2.

In Figure 1a, we illustrate the duration of audio segments and their corresponding occurrence frequencies. On average, the audio length spans 6.36 seconds, with extremes reaching 63 seconds at maximum and 1.26 seconds at minimum duration. Furthermore, Figure 1b delineates the distribution of number of characters per sentence. On average, each sentence comprises of 46 characters, with variations observed across the dataset. The dataset have a rich vocabulary of 64,082 unique words, ranging from single-character words to those with 59 characters. These metrics provide insight into the dataset's linguistic diversity and structural composition.

---

[6]Data collected from the organization *Veda Prasara Samithi*, available at https://archive.org/details/atharvaveda_202107

[7]Data collected from the organization *Veda Prasara Samithi*, available at https://archive.org/details/RigvedaChanting

[8]https://github.com/jiaaro/pydub

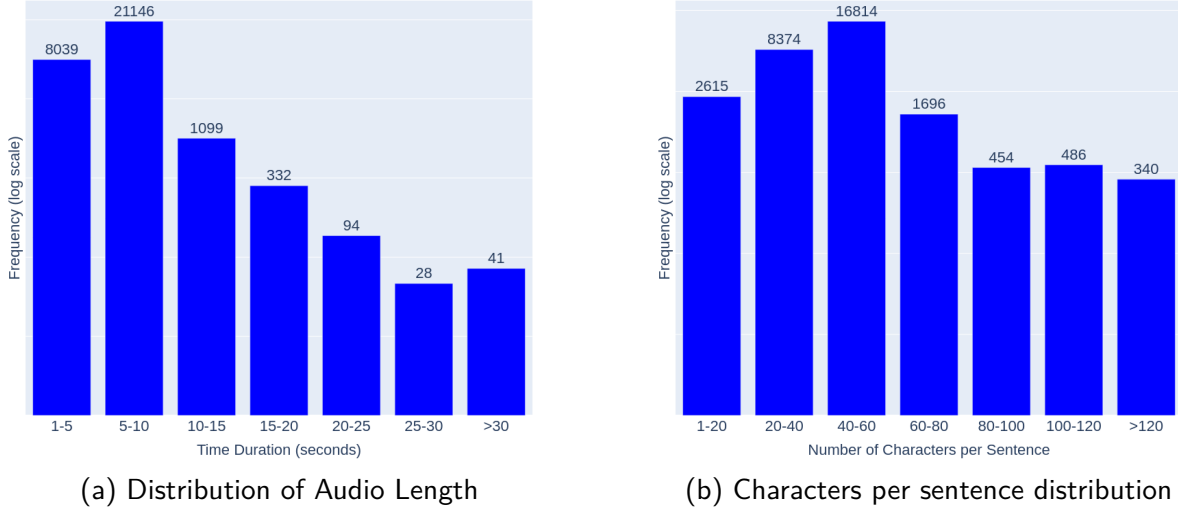

(a) Distribution of Audio Length  (b) Characters per sentence distribution

Figure 1: Statistical analysis of Sanskrit audio and text data, including the distribution of audio duration, and number of characters per sentence. All frequencies are presented on a logarithmic scale.

## 3 Experiments and Results

We explored various state-of-the-art encoder-only and encoder-decoder models. For encoder-only models, we utilized XLSR (Conneau et al., 2020) (a multilingual variant of Wav2Vec2), which employs self-supervised learning techniques to extract meaningful features from raw audio data, eliminating the need for extensive labeled datasets. Notably, XLSR is pre-trained on 436,000 hours of unlabeled speech spanning 138 languages. Additionally, we fine-tuned the MMS model (Pratap et al., 2024), which is pre-trained using Wav2Vec2's self-supervised training objective on approximately 500,000 hours of speech data in over 1,400 languages. We also fine-tuned the HuBERT model (Hsu et al., 2021), which is trained primarily on large amounts of unlabeled speech data, including the LibriSpeech dataset, which contains 960 hours of labeled audio, and the LibriLight dataset, which comprises 60,000 hours of unlabeled audio. We also fine-tuned pretrained Indic model - IndicWav2vec (Javed et al., 2022). We further carried out fine-tuning of the models on *Vedavani*, which were already fine-tuned on Sanskrit, namely SPRING-INX-MMS (Mary and Umesh, 2024), SPRING-INX-wav2vec2 (Mary and Umesh, 2024), and SPRING-INX-ccc-wav2vec2 (Mary and Umesh, 2024). Encoder-only models like Wav2Vec2 generate predictions solely based on the acoustic input, which may include noise. Consequently, the output often contains spelling mistakes as the model struggles to differentiate between homophones. To mitigate this issue, we employed n-gram language modeling using KenLM (Heafield, 2011) to construct the language model, which leverages statistical relationships between words based on their previous occurrences in the training corpus.

In addition to encoder-only models, we also explored various encoder-decoder models, specifically different variations of Whisper (Radford et al., 2023). We fine-tuned three variations of Whisper: small (244M), medium (769M), and large (1550M). The Whisper model was pre-trained using a large-scale, weakly supervised approach, leveraging 680,000 hours of paired audio and transcript data, spanning over 96 languages on various tasks, including transcription, translation, and speech recognition, focusing on multilingual and multitask learning. We also investigated distilled version of whisper i.e. Distil-Whisper (Gandhi et al., 2023) (756M). Further-more, we fine-tuned IndicWhisper (Bhogale et al., 2023b), which was developed by fine-tuning the Whisper medium model on the Vistaar (Bhogale et al., 2023b) dataset, which encompasses over 10,700 hours of audio data across 12 Indian languages, including 207 hours of Sanskrit.

Table 3: Comparison of encoder-only models for Sanskrit speech recognition, both with and without external language models. W2V refers to Wav2Vec2, and SPG-INXS represents the Sanskrit model fine-tuned by Spring Lab.

| Models | Without LM | | With LM | |
|---|---|---|---|---|
| | **WER** | **CER** | **WER** | **CER** |
| XLSR (0.3B) | 40.60 | 6.52 | 31.49 | 5.27 |
| XLSR (1B) | 38.87 | 6.38 | 31.47 | 5.36 |
| MMS (0.3B) | 63.53 | 11.46 | 45.46 | 8.67 |
| MMS (1B) | 43.93 | 7.57 | 35.38 | 6.39 |
| HuBERT | 49.63 | 8.28 | 37.83 | 6.73 |
| IndicW2V | 52.37 | 9.16 | 45.46 | 8.67 |
| SPG-INXS-MMS | 37.32 | 6.16 | **29.42** | 5.09 |
| SPG-INXS-CCC-W2V | 45.69 | 7.59 | 41.53 | 7.06 |
| SPG-INXS-W2V | **37.16** | **5.96** | 31.18 | **5.06** |

Table 4: Performance comparison of encoder-decoder models trained on IAST and Devanagari scripts, where S, M, and L denote small, medium, and large model sizes, respectively.

| Models | IAST | | Devanagri | |
|---|---|---|---|---|
| | **WER** | **CER** | **WER** | **CER** |
| Whisper-S | 35.96 | 6.35 | 28.77 | 5.26 |
| Whisper-M | 28.15 | 5.30 | 22.39 | 3.96 |
| Whisper-L | 26.05 | 4.55 | **20.71** | 3.84 |
| IndicWhisper | **23.14** | **4.12** | 21.76 | **3.81** |
| Distil-Whisper-L | 25.76 | 4.52 | 25.28 | 4.44 |

During preprocessing, we removed audio files exceeding 30 seconds to ensure a fair comparison, as Whisper does not support longer files. Our final train, development, and test datasets contained 24,590, 3,073, and 3,075 verses, respectively. In terms of the encoder-only model, we utilized the Devanagari script and identified a total of 70 unique characters, given its character-level nature. On the other hand, the encoder-decoder model employed the International Alphabet of Sanskrit Transliteration (IAST) as well as the Devanagri script. We utilized the indic-transliteration tool[9] to transliterate from Devanagri to IAST script. For a fair comparison, we have utilized SLP1 (Scharf and Hyman, 2011) script while evaluating all the models because each Sanskrit sound is represented by a single symbol in this encoding. For evaluation, we employed two standard metrics: word error rate (WER) and character error rate (CER).

### 3.1 Results Analysis

Table 3 provides a comparative analysis of the performance of various encoder-only models. The XLSR models (0.3B and 1B) achieved WER of 40.60 and 38.87, respectively. With the addition of a language model (LM), their WER improved to 31.49 and 31.47. The MMS (0.3B) model performed the worst, with a WER of 63.53, while MMS (1B) achieved a WER of 43.93. The HuBERT model showed a WER of 49.63 and a Character Error Rate (CER) of 8.28. The poor performance of IndicWav2Vec is likely due to the minimal inclusion of Sanskrit data during its pre-training phase. Both SPG-INXS-MMS and SPG-INXS-Wav2Vec models yielded similar WERs of 37.32 and 37.16 without using an LM. However, with the LM, SPG-INXS-MMS outperformed SPG-INXS-Wav2Vec by 1.76 WER points. SPG-INXS-CCC-Wav2Vec, despite being fine-tuned on a Sanskrit dataset, performed poorly with a WER of 45.69 and a CER of 7.59. Overall, we observed that using an n-gram LM improved WER by an average of 8.87 points and CER by 1.19 points across the models.

---
[9]https://github.com/indic-transliteration/indic_transliteration_py

Table 5: Distribution of word error rate (WER) and character error rate (CER) ranges, showing the count of samples falling within each range.

| WER Range | Count | CER Range | Count |
|---|---|---|---|
| 0-20 | 1837 | 0-4 | 2098 |
| 20-40 | 666 | 4-8 | 537 |
| 40-60 | 328 | 8-12 | 237 |
| 60-80 | 142 | 12-16 | 101 |
| 80-100 | 83 | 16-20 | 44 |
| $\geq 100$ | 19 | $\geq 20$ | 58 |

The results of the encoder-decoder models are presented in Table 4. The Whisper model shows improved performance with increasing model size, achieving WER of 35.96, 28.15, and 26.05 for the small, medium, and large models, respectively, on the IAST script. Additionally, the model performs better on the Devanagari script, with WER of 28.88, 22.39, and 20.71 for the small, medium, and large models, respectively. On the IAST script, IndicWhisper performs best with a WER of 23.14 and a CER of 4.12. For the Devanagari script, IndicWhisper leads in CER, but Whisper large performs slightly better in WER, with just a 0.05 difference. It was surprising to find that Distil-Whisper (756M), despite having nearly half the parameters of Whisper large (1550M), performed better on the IAST script with a WER of 25.76. However, when fine-tuned on the Devanagari script, its WER was 4.57 points higher than that of Whisper large. On average, all variants of the Whisper model performed 4.03 points better in WER and 0.7 points better in CER when fine-tuned on the Devanagari script compared to IAST.

## 4 Error Analysis

The WER and CER distributions of the generated output is presented in Table 5. As observed in the table, ∼60% samples have a WER of less than 20 while ∼68% of the samples have a CER of less than 4. However, it's important to note that a small number of samples have WER values exceeding 100. This high WER is attributed to word-splitting issues, where the predictions contain more words than the reference. This discrepancy suggests that while the models perform well on average, there are specific challenges related to word boundary detection and segmentation that need to be addressed.

Additionally, the distribution shows that fewer samples fall into higher WER ranges. Similarly, for CER, the majority of samples fall within the lower range, which is a positive indicator of the models' performance. The presence of higher CER values for a small fraction of samples indicates that there might be specific phonetic or orthographic challenges that could be explored further. Improving handling of word boundaries and refining phoneme recognition could potentially enhance the accuracy of the models, especially for cases with higher WER and CER. Future work could focus on addressing these specific error types to improve the overall performance and reliability of the ASR systems.

In Table 6, we present examples of ASR-generated output using the best-performing model and compare it with the reference transcriptions. These examples highlight different types of errors encountered in Vedic Sanskrit Speech Recognition, including phonetic, structural, and intonational discrepancies. Phonetic errors are evident in cases where similar consonants (e.g., श, ष, स) or short-long vowels are interchanged, leading to incorrect predictions. Structural errors include word-splitting, merging, and Sandhi misrecognition, which significantly impact meaning and syntactic integrity.

Table 6: Examples of sentences with different error types in Vedic Sanskrit Speech Recognition: Phonetic, Structural, and Intonational

| Reference | Predicted |
|---|---|
| रथं न चित्रं वपुषाय दर्शतं मनुर्हितं सदमिद्राय ईमहे | रथं न चित्रं वपुषाय दर्शतं मनुर्हितं सदमिद्राय ईमहे |
| वनस्पतीन् वानस्पत्यान् ओषधीरुत वीरुधः | वनस्पतीन् वानस्पत्यान् ओषधीरुत विरुधः |
| तेषामिष्टानि विहितानि धामश स्थात्रे रेजन्ते विकृतानि रूपशः | तेषामिष्टानि विहितानि धामश स्थात्रे रेजन्ते विहितानि रूपशः |
| महि महे तवसे दीध्ये नॄनिन्द्रायेत्था तवसे अतव्यान् | महि महे तवसे दीध्ये निरिन्द्रायेऽत्था तवसे अतव्यान् |
| घृतं मिमिक्षे घृतमस्य योनिर्घृते श्रितो घृतम्वस्य धाम | घृतं मिमिक्षे घृतमस्ययोनिर्घृते श्रुतो घृतं वस्य धाम |
| ततः किशोरा म्रियन्ते वत्सांश्च घातुको वृकः | ततः किशोरा म्रियन्ते वत्सांश्च गातु को वृकः |
| यत्कृषते यद्धनुते यच्च वक्षेन विन्दते सर्व मर्त्यस्य तन् नास्ति क्रव्याच्चेदनिराहितः | यत्कृषते यद्धरुते यच्च वक्षेन विन्दते सर्व मर्त्यस्य तन् नास्ति क्रव्याचिदनिराहितः |
| तस्मै ते नक्षत्रराज शकधूम सदा नमः | तस्मै ते नक्षत्त राजस्य कधूम सदा नमः |

Additionally, in Table 7, we present a range of errors encountered in ASR outputs, categorized based on their linguistic nature. One prominent class of errors includes word-splitting and word-conjunction (sandhi errors), where words are either incorrectly segmented or merged inappropriately. For instance, the phrase ब्रह्मन्नविष was erroneously split into ब्रह्मं नविष, whereas जातवेदः स्याम was incorrectly merged into जातवेदस्याम. These errors disrupt syntactic structure and often alter the meaning of the text. Additionally, phoneme substitution errors were observed, particularly in cases where similar phonetic units were misrecognized. This includes errors in aspirated and unaspirated consonants, as seen in कब्द्ध रथ, which was incorrectly predicted as कब्द्रस्त, where थ was replaced by त. Similarly, confusion between phonetically close consonants like श, स, and ष was frequent, leading to errors such as वृजिनानामविष्यवः being predicted as वृजिनानामभिश्यवः.

Furthermore, nasalization errors were prevalent, where different classes of nasal phonemes were misrepresented. For example, the guttural nasal ङ was incorrectly replaced with the dental nasal न, affecting phonetic accuracy and pronunciation consistency. In addition to phoneme-level errors, character-level errors such as insertions, omissions, and substitutions were also common. For instance, तृतीया was incorrectly predicted as तृतीयाः due to unnecessary vowel elongation, and त्य्ये was reduced to त्वे, omitting a critical character. These errors highlight the inherent difficulties in processing Sanskrit's rich phonological system, morphological complexity, and intricate Sandhi transformations. Addressing these challenges requires ASR models to incorporate linguistic priors, improved phoneme modeling, and enhanced contextual awareness to achieve more accurate transcriptions.

## 5 Conclusion & Future Work

We have introduced a new dataset for Vedic ASR that complements the existing dataset to broaden the spectrum of ASR for Sanskrit. In addition to the dataset, we also performed benchmarking of the ASR models on our newly introduced dataset and studied the performance of these models, thereby setting up the stepping stone for future research work.

The Vedas are classified as Shrutis, meaning *that which is heard*. As such, their transcription includes precise indications of intonation, which are typically standard, low, high, and high-high (a high tone followed by another high tone). Accurately capturing these intonations is crucial for improving the model's ability to recognize and differentiate the characters associated with each intonation. In future work, we plan to enhance the dataset's transcripts by including detailed markers for intonation. This enhancement will enable the development of more robust models or the creation of new models specifically designed to recognize and process these intonational features effectively.

Table 7: Types of errors observed in the ASR output, categorized by the nature of the discrepancy between the reference and predicted texts. Comments explain the change in each set of errors.

| Error Type | Comments | Reference | Predicted |
|---|---|---|---|
| Word-Split | संधि विच्छेद | ब्रह्मन्नविष | ब्रह्मं नविष |
| Word-Conjuction | संधि | जातवेदः स्याम | जातवेदस्याम |
| Wrong Prediction | ध -> प | धावमानं | पावमानं |
| | स -> म | संसृजति | सुमृजति |
| | ज -> य | जेषाम | येषाम |
| | अनुमान दोषः | अण्व्या | अन्या |
| | ढ -> ध | दृढान् | दृधान् |
| Phoneme Substitution | ष -> श | वृजिनानामविष्यवः | वृजिनानामभिश्यवः |
| | स -> श | व्यख्यज्जिह्वयासितः | व्यख्यज्जिह्वयाशितः |
| | श -> स | सूर्यरश्मिर्हरिकेशः | सूर्यदस्मिर्हरिकेशः |
| | ये -> ए | येषां | एषां |
| Unaspirated-Aspirated | थ -> त | कद्रु स्थ | कद्रुस्त |
| | ख -> क | सखिवाँ | सकिवाँ |
| Nasal | Guttural ङ -> म | साकमधराङ्रेहि | साकमधरां परेहि |
| | Palatal ञ -> ञ | जुजुर्वान्दशमे | जुजुर्वाञ्दशमे |
| | Retroflex ण -> न | पण्श्रिद्धि म्रदा | पनेश्रिद्धिन्द्रदा |
| | Dental न -> म | नही | मही |
| | Labial म -> व | पितृतमः | पितृतवः |
| Character Insertion | विसर्ग आगम | तृतीया | तृतीयाः |
| Character Omission | य लोपः | त्व्ये | त्वे |
| Character Substitution | र -> न | अरंकृतः | अनंकृतः |

## Acknowledgements

We appreciate and thank all the anonymous reviewers for their constructive feedback towards improving this work. The work was supported in part by the National Language Translation Mission (NLTM): Bhashini project by the Government of India.

## References

[Adiga et al.2021] Devaraja Adiga, Rishabh Kumar, Amrith Krishna, Preethi Jyothi, Ganesh Ramakrishnan, and Pawan Goyal. 2021. Automatic speech recognition in Sanskrit: A new speech corpus and modelling insights. Online, August. Association for Computational Linguistics.

[Anoop and Ramakrishnan2019] CS Anoop and AG Ramakrishnan. 2019. Automatic speech recognition for sanskrit. In *2019 2nd International Conference on Intelligent Computing, Instrumentation and Control Technologies (ICICICT)*, volume 1, pages 1146–1151. IEEE.

[Bhogale et al.2023a] Kaushal Bhogale, Abhigyan Raman, Tahir Javed, Sumanth Doddapaneni, Anoop Kunchukuttan, Pratyush Kumar, and Mitesh M Khapra. 2023a. Effectiveness of mining audio and text pairs from public data for improving asr systems for low-resource languages. In *Icassp 2023-2023 ieee international conference on acoustics, speech and signal processing (icassp)*, pages 1–5. IEEE.

[Bhogale et al.2023b] Kaushal Santosh Bhogale, Sai Sundaresan, Abhigyan Raman, Tahir Javed, Mitesh M Khapra, and Pratyush Kumar. 2023b. Vistaar: Diverse benchmarks and training sets for indian language asr. *arXiv preprint arXiv:2305.15386*.

[Conneau et al.2020] Alexis Conneau, Alexei Baevski, Ronan Collobert, Abdelrahman Mohamed, and Michael Auli. 2020. Unsupervised cross-lingual representation learning for speech recognition. *arXiv preprint arXiv:2006.13979*.

[Gandhi et al.2023] Sanchit Gandhi, Patrick von Platen, and Alexander M Rush. 2023. Distil-whisper: Robust knowledge distillation via large-scale pseudo labelling. *arXiv preprint arXiv:2311.00430*.

[Heafield2011] Kenneth Heafield. 2011. KenLM: Faster and smaller language model queries. In *Proceedings of the Sixth Workshop on Statistical Machine Translation*, pages 187–197, Edinburgh, Scotland, July. Association for Computational Linguistics.

[Hsu et al.2021] Wei-Ning Hsu, Benjamin Bolte, Yao-Hung Hubert Tsai, Kushal Lakhotia, Ruslan Salakhutdinov, and Abdelrahman Mohamed. 2021. Hubert: Self-supervised speech representation learning by masked prediction of hidden units. *IEEE/ACM transactions on audio, speech, and language processing*, 29:3451–3460.

[Javed et al.2022] Tahir Javed, Sumanth Doddapaneni, Abhigyan Raman, Kaushal Santosh Bhogale, Gowtham Ramesh, Anoop Kunchukuttan, Pratyush Kumar, and Mitesh M Khapra. 2022. Towards building asr systems for the next billion users. In *Proceedings of the AAAI Conference on Artificial Intelligence*, volume 36, pages 10813–10821.

[Javed et al.2023] Tahir Javed, Kaushal Bhogale, Abhigyan Raman, Pratyush Kumar, Anoop Kunchukuttan, and Mitesh M Khapra. 2023. Indicsuperb: A speech processing universal performance benchmark for indian languages. In *Proceedings of the AAAI Conference on Artificial Intelligence*, volume 37, pages 12942–12950.

[Krishna et al.2018] Amrith Krishna, Bishal Santra, Sasi Prasanth Bandaru, Gaurav Sahu, Vishnu Dutt Sharma, Pavankumar Satuluri, and Pawan Goyal. 2018. Free as in free word order: An energy based model for word segmentation and morphological tagging in sanskrit.

[Krishna et al.2021] Amrith Krishna, Bishal Santra, Ashim Gupta, Pavankumar Satuluri, and Pawan Goyal. 2021. A Graph-Based Framework for Structured Prediction Tasks in Sanskrit. *Computational Linguistics*, 46(4):785–845, 02.

[Lal2015] B. B. Lal. 2015. *The Rigvedic people 'invaders'?/'immigrants'? or indigenous? : evidence of archaeology and literature.* Aryan Books International, New Delhi, India.

[Mary and Umesh2024] NJ Metilda Sagaya Mary and S Umesh. 2024. Inx-speakerhub: A 2000-hour indian multiligual speaker verification corpus. In *2024 IEEE Spoken Language Technology Workshop (SLT)*, pages 1217–1223. IEEE.

[Pratap et al.2024] Vineel Pratap, Andros Tjandra, Bowen Shi, Paden Tomasello, Arun Babu, Sayani Kundu, Ali Elkahky, Zhaoheng Ni, Apoorv Vyas, Maryam Fazel-Zarandi, et al. 2024. Scaling speech technology to 1,000+ languages. *Journal of Machine Learning Research*, 25(97):1–52.

[Radford et al.2023] Alec Radford, Jong Wook Kim, Tao Xu, Greg Brockman, Christine McLeavey, and Ilya Sutskever. 2023. Robust speech recognition via large-scale weak supervision. In *International Conference on Machine Learning*, pages 28492–28518. PMLR.

[Scharf and Hyman2011] Peter M Scharf and Malcolm D Hyman. 2011. Linguistic issues in encoding sanskrit. *The Sanskrit Library*, 20.

[Suhani et al.2023] Suhani, Amita Dev, and Poonam Bansal. 2023. Ctc-based end-to-end speech recognition for low resource language sanskrit. In *2023 26th Conference of the Oriental COCOSDA International Committee for the Co-ordination and Standardisation of Speech Databases and Assessment Techniques (O-COCOSDA)*, pages 1–5.

[Wright1968] J. C. Wright. 1968. J. f. staal: Word order in sanskrit and universal grammar. (foundations of language supplementary series, vol. 5.) xi, 98 pp. dordrecht: D. reidel publishing co., 1967. guilders 30. *Bulletin of the School of Oriental and African Studies*, 31(1):205–205.